\documentclass[conference, onecolumn]{IEEEtran}
\usepackage{cite}
\usepackage{amsmath,amssymb,amsfonts,mathrsfs,mathtools}
\usepackage{graphicx}
\usepackage{textcomp}
\usepackage{xcolor}
\usepackage{algorithm}
\usepackage{algpseudocode}
\usepackage{amsmath}
\usepackage{amssymb}
\usepackage{caption}
\usepackage{subcaption}
\usepackage{multirow} 

\usepackage{algorithmicx}
\usepackage{algpseudocode}

\usepackage{xcolor}
\usepackage{listings}

\usepackage{tabularx}
\newcolumntype{L}{>{\centering \arraybackslash}X}

\definecolor{keywordcolor}{rgb}{0.0,0.0,0.6}
\definecolor{commentcolor}{rgb}{0.0,0.5,0.0}
\definecolor{stringcolor}{rgb}{0.58,0,0.82}

\def\BibTeX{{\rm B\kern-.05em{\sc i\kern-.025em b}\kern-.08em
    T\kern-.1667em\lower.7ex\hbox{E}\kern-.125emX}}

\newcommand{\un}[1]{\underline{#1}}

\newtheorem{defn}{Definition}
\newtheorem{thm}{{\cal T}heorem}
\newtheorem{cor}{Corollary}
\newtheorem{prop}{Proposition}
\newtheorem{lem}{Lemma}
\newtheorem{conj}{Conjecture}
\newtheorem{constr}{Construction}
\newtheorem{note}{Remark}

\newcommand{\bit}{\begin{itemize}}
	\newcommand{\eit}{\end{itemize}}
\newcommand{\bcor}{\begin{cor}}
	\newcommand{\ecor}{\end{cor}}
\newcommand{\beq}{\begin{equation}}
	\newcommand{\eeq}{\end{equation}}
\newcommand{\beqn}{\begin{equation}}
	\newcommand{\eeqn}{\end{equation}}
\newcommand{\bea}{\begin{eqnarray}}
	\newcommand{\eea}{\end{eqnarray}}
\newcommand{\bean}{\begin{eqnarray*}}
	\newcommand{\eean}{\end{eqnarray*}}
\newcommand{\ben}{\begin{enumerate}}
	\newcommand{\een}{\end{enumerate}}
\newcommand{\bdefn}{\begin{defn}}
	\newcommand{\edefn}{\end{defn}}
\newcommand{\bnote}{\begin{note}}
	\newcommand{\enote}{\end{note}}
\newcommand{\bprop}{\begin{prop}}
	\newcommand{\eprop}{\end{prop}}
\newcommand{\blem}{\begin{lem}}
	\newcommand{\elem}{\end{lem}}
\newcommand{\bthm}{\begin{thm}}
	\newcommand{\ethm}{\end{thm}}
\newcommand{\bconj}{\begin{conj}}
	\newcommand{\econj}{\end{conj}}
\newcommand{\bconstr}{\begin{constr}}
	\newcommand{\econstr}{\end{constr}}
\newcommand{\bpf}{\begin{proof}}
	\newcommand{\epf}{\end{proof}}
\newcommand{\bprf}{{\em Proof: }}
\newcommand{\eprf}{\hfill $\Box$}

\newcommand{\und}[1]{\underline{#1}}

\IEEEoverridecommandlockouts

\begin{document}

\title{Link Adaptation Using Joint-Thompson Sampling}

\author{
\IEEEauthorblockN{Vignatha~Vinjam\IEEEauthorrefmark{1} \quad Manjunath Kolavennu\IEEEauthorrefmark{1}\quad Myna~Vajha\IEEEauthorrefmark{1} \quad Karthik Periyapattana Narayanaprasad\IEEEauthorrefmark{2}}
\IEEEauthorblockA{\IEEEauthorrefmark{1} Department of EE, IIT Hyderabad \qquad \IEEEauthorrefmark{2} Department of AI, IIT Hyderabad\\
{ ee23resch14002@iith.ac.in, ee25mtech02002@iith.ac.in, mynav@ee.iith.ac.in, pnkarthik@ai .iith.ac.in}}
\thanks{Myna Vajha is supported by ANRF-ECRG and IIT Hyderabad Seed Grant.}
}

% \iffalse
% \author{\IEEEauthorblockN{ Vignatha Vinjam}
% \IEEEauthorblockA{\textit{EE, IITH}
% email}
% \and
% \IEEEauthorblockN{Manjunath Kolavennu}
% \IEEEauthorblockA{\textit{EE, IITH}
% {\color{red} please add your email, check your name }}
% \and
% \IEEEauthorblockN{ Myna Vajha}
% \IEEEauthorblockA{\textit{EE, IITH} \\
% mynav@ee.iith.ac.in}
% \and
% \IEEEauthorblockN{ Karthik P N}
% \IEEEauthorblockA{\textit{EE, IITH} \\
% pnkarthik@ai.iith.ac.in}
% }
% \fi 

\maketitle

\begin{abstract}
The choice of Modulation and Coding (MCS) type for a particular channel condition is made through link adaptation (LA) algorithms that operate at the MAC layer. These algorithms rely on the ACK/NACK statistics and the channel quality index (CQI) feedback. Several existing works model LA as a multi-armed bandit (MAB) problem across cellular and Wi-Fi links. In the MAB formulation, each available MCS is a Bernoulli arm parameterized by its transmission success probability, and the goal is to design a selection strategy that accrues maximum reward. Several popular MAB algorithms, such as upper confidence bound (UCB) and Thompson Sampling (TS), have been proposed in the literature. Using the fact that MCS success probabilities are ordered, we propose the Joint-Thompson Sampling (Joint-TS) algorithm. Unlike classical TS, which assumes independent Beta distributions for each arm, Joint-TS utilizes a multivariate ordered Beta distribution as the prior to preserve the inherent monotonicity of success probabilities. Our simulation results show that while existing MAB algorithms fail in specific scenarios, Joint-TS delivers competitive throughput with robust, consistent performance in all scenarios.
\end{abstract}


\section{Introduction}
Link adaptation (LA) is an essential MAC layer algorithm that selects link parameters such as Modulation and Coding Scheme (MCS), power control, number of layers, and precoders for dynamic wireless channels. LA focusing on MCS selection is widely known as Adaptive Modulation and Coding (AMC) or as Adaptive Data Rate (ADR) in WiFi. 
%. 
The goal of MCS selection is to maximize expected throughput under an error rate constraint for a given channel state---the Signal-to-Interference-plus-Noise Ratio (SINR). Mathematically,
\bean
R_{\rm max} (c) &=& \max_{i: ~ \mathrm{BLER}(i, c) \, \le \, p_{\text{target}}} s_i \cdot (1-\mathrm{BLER}(i,c)),
\eean
where $s_i$ is the spectral efficiency corresponding to MCS~$i$, determined as product of number of bits per modulation symbol times the rate of forward error correction (FEC) code  used
%{\color{blue} (Karthik: can we mention the full-form of FEC here, and continue to use the abbreviation FEC elsewhere?)}
(cf. Table~\ref{tab:MCS_Table_5G}), and $\mathrm{BLER}(i, c)$ is the Block Error Rate at channel state (or effective SINR)~$c$ 
 %({\color{red} cite effective SINR})
upon using MCS~$i$. Effective SINR \cite{eff_SINR} maps the sub-carrier SINR vector to a single scalar, enabling the use of Additive White Gaussian Noise (AWGN) BLER tables. While the BLER target, $p_{\text{target}}$, is typically set to $0.1$ \cite{bler_target}, SINR-dependent BLER targets result in better performance of ARQ schemes at the MAC layer.\cite{optimal_target_bler}  If the throughput-maximizing MCS always satisfies this BLER constraint, then the condition ${\rm BLER}(i,c) \leq p_{\rm target}$ can be ignored. We observe this holds true for the 5G NR waveform simulations conducted over the AWGN channel using the MATLAB 5G toolbox. 
%{\color{red} please check this line.}
% \\Under the condition
%     \begin{equation*}
%         {BLER}(k,s) \le BLER_{\text{target}} = 0.1 \text{ (typically)}
%     \end{equation*}
\begin{table}[!th]
\centering
\scalebox{0.7}{\begin{tabular}{|c|c|c|c|}
\hline
\textbf{MCS $(i)$ } & $M_i$ & $F_i$ & $s_i$ \\
\hline
0  & 2 & 120 & 0.2344 \\
1  & 2 & 157 & 0.3066 \\
2  & 2 & 193 & 0.3770 \\
3  & 2 & 251 & 0.4902 \\
% 4  & 2 & 308 & 0.6016 \\
% 5  & 2 & 379 & 0.7402 \\
\vdots & \vdots & 
\multicolumn{1}{c|}{\vdots} & \vdots \\
% 6  & 2 & 449 & 0.8770 \\
% 7  & 2 & 526 & 1.0273 \\
% 8  & 2 & 602 & 1.1758 \\
% 9  & 2 & 679 & 1.3262 \\
% 10 & 4 & 340 & 1.3281 \\
% 11 & 4 & 378 & 1.4766 \\
% 12 & 4 & 434 & 1.6953 \\
% 13 & 4 & 490 & 1.9141 \\
% 14 & 4 & 553 & 2.1602 \\
% 15 & 4 & 616 & 2.4063 \\
% 16 & 4 & 658 & 2.5703 \\
% 17 & 6 & 438 & 2.5664 \\
% 18 & 6 & 466 & 2.7305 \\
% 19 & 6 & 517 & 3.0293 \\
% 20 & 6 & 567 & 3.3223 \\
% 21 & 6 & 616 & 3.6094 \\
% 22 & 6 & 666 & 3.9023 \\
% 23 & 6 & 719 & 4.2129 \\
% 24 & 6 & 772 & 4.5234 \\
% 25 & 6 & 822 & 4.8164 \\
% 26 & 6 & 873 & 5.1152 \\
27 & 6 & 910 & 5.3320 \\
28 & 6 & 948 & 5.5547 \\
\hline
\end{tabular}}
\caption{5G NR MCS configurations\cite[5.1.3.1 Table~1]{3gpp38_214}, where $M_i$ is bits/symbol, $F_i$ is FEC code rate ($\times 1024$), and $s_i = M_i \cdot F_i/1024$ is the spectral efficiency.
} 
\label{tab:MCS_Table_5G}
\end{table}

While uplink MCS selection is straightforward using the SINR estimates, the downlink selection is a complex task since the gNB lacks direct channel knowledge. As a remedy, a quantized MCS suggestion, called Channel Quality Index (CQI), can be requested from the UE. See \cite[ 5.2.2.1]{3gpp38_214} for a list of 5G NR CQI options.
\iffalse
\begin{table}[ht!]
(\scalebox{0.85}{
\begin{tabular}{cccccccccccccccc}
     0& 1& 2& 4& 6& 8& 11& 13& 15& 18& 20& 22& 24& 26& 28
\end{tabular}
})
\caption{MCS values present in the 5G NR CQI table ( see \cite{3gpp38_214}, Table 5.2.2.1-2)\label{tab:cqi_list}} %{\color{red} Vignatha, please check this table for correctness and add citation, why do I see 20 here ?}}
\end{table}
\fi

% The traditional link adaptation algorithm uses a combination of offline generated CQI-to-MCS lookup table and an error-rate based feedback loop. However, this approach can be limited by link modeling errors and slow convergence, especially in dynamic channel conditions. Recent works have explored reinforcement learning-based LA as alternatives which have shown promise for faster adaptation and increased robustness. 

\vspace{4pt}
\paragraph{Prior Work}
% {\color{red}Describe the literature that introduced OLLA and ILLA and motivate why there was need to model using multi-armed bandits, refer to Vidit's thesis. Please find, how ordering is used in Vidit's work to decrease the number of samples used}
Link adaptation in wireless systems has traditionally relied on two core mechanisms: Inner Loop Link Adaptation (ILLA) and Outer Loop Link Adaptation (OLLA) (see \cite{OLLA_3G}, \cite{OLLA_LTE}). ILLA maps CQI or estimated SINR to an MCS using predefined lookup tables, while OLLA works in conjunction by adding an SINR offset adjusted via Acknowledgment (ACK)/ Negative Acknowledgment (NACK) feedback. The ratio of upward to downward offset adjustments is determined by the target BLER, ensuring the expected BLER remains constrained. Although OLLA is effective, it suffers from slow responses to fast-changing channels\cite{OLLA_shortcomings}, sensitivity to step sizes, and a lack of per-CQI state information.
An optimization scheme for initial OLLA offset selection is proposed in \cite{OLLA_self_opt}; however, this requires periodic intervention at each cell-site.
To address these limitations, recent literature models LA as a multi-armed bandit (MAB) problem. MAB framework models each MCS as an arm generating Bernoulli rewards, allowing the algorithm to learn from history of rewards and arm selections in an online fashion. This facilitates principled exploration of uncertain arms while ensuring performance is maximized, as more reward samples are observed. MAB-based solutions can outperform OLLA in highly dynamic or unpredictable channels, requiring no manual parameter tuning and adapting quickly based on observed feedback. In \cite{UCB}, the authors formulate rate adaptation in 802.11 as an MAB problem and propose a KL-UCB based algorithm called G-ORS, which assumes unimodality of the expected throughput. $\epsilon$-greedy approaches \cite{vidit_epsilon} balance random exploration with the greedy exploitation of the empirically best-performing arm. MTS \cite{MTS} and BayesLA \cite{Vidit_TS} have similar approaches based on Thompson Sampling (TS). Here, independent Beta priors are maintained for the success probability of each MCS, at every CQI. In \cite{UTS}, a TS-based alternative to G-ORS is proposed. In \cite{Vidit_LatentTS} and \cite{wifi}, a version of the TS algorithm that operates on channel SINR values is proposed, wherein SINR estimates are mapped to MCS values using a lookup table along similar lines as in OLLA.

% {\color{red} describe UCB, Modified TS, Unimodal TS, Latent TS, CoTS(Srikant), cover the  Russian paper that uses latentTS.}
%For the remainder of this paper, we will refer to the Traditional approach as OLLA and MTS as TS. 
% {\color{red} take care of latentTS citation too}
\paragraph{Our Contributions}
We propose Joint-TS, a version of classical TS that exploits the monotonicity property of success probabilities across MCS values (i.e., lower MCS yields higher success probability). Unlike the independent Beta distributions employed in classical TS, Joint-TS utilizes a multivariate ordered Beta (MOB) prior \cite{OrderedBeta_SaidiKuznetsovNediak}. The advantages of employing MOB distribution are: (a) the samples satisfy the monotonicity constraint, and (b) the posterior remains an MOB distribution. To address the challenge of sampling from the MOB distribution, we propose a method based on Gibbs sampling, a popular MCMC technique.

Using the pyitpp library \cite{py_itpp}, we demonstrate that while state-of-the-art unimodal algorithms (like UTS \cite{UTS}) perform well with perfect CQI, their performance degrades significantly when CQI feedback is unavailable. Joint-TS maintains stable throughput in these deprived conditions. Furthermore, we highlight the limitations of Latent-TS (LTS) \cite{Vidit_LatentTS}, which relies heavily on a pre-existing lookup table. While LTS performs well ideally, it is vulnerable in high Doppler environments and practical setups with inaccurate tables. Joint-TS operates independently of such tables while matching or exceeding LTS performance in high-mobility states.

The idea of exploiting the monotonicity property of the success probabilities appears in \cite{conTS}, wherein the authors propose Constrained TS (CoTS), an algorithm in which sampling from an MOB prior is achieved using a sequential inverse transform sampling technique. However, their technique results in samples that are not, in fact, jointly MOB distributed; for a more detailed discussion, see Sec.~\ref{sec:sampling_mob}. Additionally, while \cite{conTS} focuses on regret analysis for CoTS, our focus is on the performance of Joint-TS for Doppler scenarios.
% \subsection{Organization of Paper}
% {\color{blue} (Karthik: It is ok to omit this section if there isn't enough space to include it).}
% We discuss the multi-armed bandit formulation of the link adaptation problem in Sec.~\ref{sec:mab} and present the Joint-Thompson Sampling (TS) algorithm  in Sec.~\ref{sec:joint_ts} that exploits the monotonicity property of the priors by sampling the priors for the arms jointly from a multi-ordered Beta (MOB) distribution. We present an efficient sampling algorithm for the MOB distribution in Sec.~\ref{sec:sampling_mob} that uses properties of the MOB distributions that we present in this paper. We present simulation results of Joint-TS algorithm and compare with TS algorithm in Sec.~\ref{sec:simulations}.

{\em Notations:} 
%$\mathbb{N}=\{1,2,\ldots\}$ denotes the set of natural numbers, and 
For any positive integer $K$, let $[K] \coloneqq \{1, 2, \cdots, K\}$. Given a vector $\und{\Theta}=(\Theta_1, \ldots, \Theta_K)$ of length $K$, $\un{\Theta}_{\sim i}$ denotes the vector $(\Theta_1, \cdots, \Theta_{i-1}, \Theta_{i+1}, \cdots, \Theta_K)$ of all elements in vector $\un{\Theta}$ except $\Theta_i$. We employ the following abbreviations: (a) i.i.d: independent and identically distributed, (b) PDF: probability density function, (c) CDF: cumulative distribution function, and (d) TTI: transmission time interval. 

%$[a, b] \coloneqq \{ x \in \mathbb{R} \mid a \le x \le b\}$. 

% \begin{figure}
%     \centering
%     \includegraphics[width=\linewidth]{traditional.pdf}
%     \caption{ Traditional (state-of-the-art) Approach with 2 control loops }
%     \label{fig:enter-label}
% \end{figure}
% {\color{red} please add .pdf version for the figures so that the quality is good. if you are using draw.io, please download the pdfs, also run the command pdfcrop on top of pdfs so that we can use space effectively}

\section{The MAB Problem Formulation}\label{sec:mab}

The MCS selection problem in link adaptation can be formulated as a multi-armed bandit (MAB) problem \cite{thompson1933likelihood}. While we assume the channel state (SINR) to be fixed but unknown in our problem description, this formulation extends to time-varying channels (i.e., Doppler scenarios), as outlined in Sec.~\ref{sec:simulations}. Let $[K]$ denote the set of $K$ supported MCS levels, each representing a distinct arm. At every discrete time step $t = 1, 2, \ldots, T$, the transmitter selects an MCS arm $I_t \in [K]$ and obtains the reward $R_{I_t, t} = s_{I_t}  X_{I_t, t}$. Here, $s_{I_t}$ is the known spectral efficiency of MCS $I_t$, and $X_{I_t,t} \sim \text{Bernoulli}(\theta_{I_t})$ represents the transmission outcome (1 for success, 0 for failure), governed by the unknown success probability $\theta_{I_t}$. The spectral efficiencies are known and satisfy the relation $s_1 < s_2 < \dots < s_K$, while the success probabilities are unknown but are known to be monotonically decreasing, satisfying $\theta_1 \ge \theta_2 \ge \dots \ge \theta_K$.

% which results in a  reward $s_i X_i$ $\mathrm{Bernoulli}(\theta_i)$ arm of the bandit, and the reward associated with each arm is the \emph{goodput} (i.e., the amount of successfully delivered data) obtained when transmitting with that MCS under prevailing channel conditions.

% {\color{red}spectral efficiency and monotonicity aspects are not coming to picture here, please use Bernoulli($\theta_i$) to describe each MCS value's success probability and reward as a scaling of this Bernoulli R.V by spectral efficiency corresponding to that MCS. Please also fix the arg max step in Joint-Thompson Sampling step to scale it with the spectrical efficiency of the MCS value}

% The stochastic reward associated with playing arm $k$ (selecting MCS $k$) is a random variable, denoted $r_{k, t}$, whose distribution depends on both the MCS and the time-varying, potentially unknown, wireless channel conditions:
% \[
% r_{k, t} \sim \mathbb{P}_k, \qquad k = 1, \ldots, K
% \]
% where $\mathbb{P}_k$ represents the (unknown) reward distribution for arm $k$ (i.e., the distribution of goodput when using MCS $k$).

The goal of the transmitter is to maximize the expected cumulative {\em reward}  (throughput) over the time horizon $T$. %, defined as
%\[
% \mathrm{Regret}(T) \coloneqq T s_{i^*} \theta_{i^*} - \mathbb{E}\left[\sum_{t=1}^T R_{I_t, t}\right],
% \]
% where $i^* = 
% %\arg\max_{i \in [K]} \mathbb{E}[R_{i, t}] = 
% \arg\max_{i \in [K]} s_i \theta_i,
% $ denotes the arm with the highest mean reward. 
This requires balancing the exploration of untested MCS values with the exploitation of the empirically best-performing ones via sequential interaction with the environment.
%The performance of transmitter's strategy is measured in comparison to the performance of the best strategy in hindsight via \emph{regret}, defined as
% the difference between the expected cumulative reward of always selecting the best arm and the cumulative reward achieved by the policy:

%The regret minimization is still non-trivial as the optimization could result in any MCS value depending on the channel condition as the exact values of $\theta_i$'s that are unknown {\color{blue} Karthik: this sentence is not very clear to me}. 
We propose Joint-TS, a version of the classical TS algorithm, that exploits and retains the monotonicity property of the success probabilities throughout the entire time horizon $T$.

% \begin{figure}
%     \centering
%     \includegraphics[width=\linewidth]{RL.pdf}
%     \caption{Multi-armed-bandit formulation of link adaptation problem}
%     \label{fig:LA-as-MAB}
% \end{figure}

\section{The Joint-Thompson Sampling Algorithm}\label{sec:joint_ts}

In the classical TS algorithm \cite{thompson1933likelihood}, the unknown success probabilities are treated as random variables $\Theta_1, \ldots, \Theta_K$, with an initial prior specified by the product Beta distribution: $\Theta_i \sim {\rm Beta}(\alpha_i, \beta_i)$  independent of $\und{\Theta}_{\sim i}$ for each $i \in [K]$. Here, $\alpha_1, \ldots, \alpha_K$ and $\beta_1, \ldots, \beta_K$ are fixed constants. At time step $t$, the TS algorithm selects arm $I_t = \mathrm{argmax}_{i \in [K]} s_i \Theta_i$. Upon obtaining reward $R_t=r$ from arm $I_t$, the parameters $(\alpha_{I_t}, \beta_{I_t})$ of arm $I_t$ are updated as in \eqref{eq:parameter-updates} (with $I_t=i$) while leaving the parameters of other arms unchanged. The independence of $\Theta_1, \ldots, \Theta_K$ assumed at the start of TS algorithm clearly fails to capture the monotonicity in the success probabilities of the arms.
%As evident, the initial product distribution in the TS algorithm fails to capture the monotonicity property of the success probabilities.
%The pseudo code corresponding to Thompson-Sampling algorithm is as provided in Algortihm~\ref{alg:ts_pc} where 

To preserve the monotonicity relation $\Theta_1 \geq \Theta_2 \geq \cdots \geq \Theta_K$ at every time step, we propose the Joint-TS algorithm in which $\Theta_1, \ldots, \Theta_K$ are sampled from a Multivariate Ordered Beta (MOB) distribution, introduced in \cite{OrderedBeta_SaidiKuznetsovNediak}. 
%This is crucial, as employing a larger MCS implicitly leads to higher success probability (throughput). Instead of sampling $\Theta_1, \ldots, \Theta_K$ from a product Beta distribution at the start, , 
Sampling from the MOB distribution inherently guarantees that the resulting samples satisfy the desired monotonicity constraint. The pseudocode for Joint-TS is presented in Algorithm~\ref{alg:jointts_pc}, and its key facets are described below.
% By virtue of sampling from MOB distribution, the resulting samples satisfy the desired monotonicity property. The pseudocode of the Joint-TS algorithm is presented in Algorithm~\ref{alg:jointts_pc}. %To the best of our knowledge, this is the first work on MAB-based link adaptation to capture the above monotonicity property. 
% Below, we describe some important facets of the Joint-TS algorithm.

\subsection{Multivariate Ordered Beta (MOB) Distribution}
Fix $\und{\alpha} = (\alpha_1, \cdots, \alpha_K)$ and $\und{\beta} = (\beta_1, \cdots, \beta_K)$ with $\alpha_i, \beta_i > 0$ for all $i \in [K]$. We say that $\und{\Theta}=(\Theta_1, \ldots, \Theta_K)$ follows MOB distribution with parameters $(\und{\alpha}, \und{\beta})$, denoted by ${\rm MOB}(\und{\alpha}, \und{\beta})$, if the joint PDF of $\Theta_1, \ldots, \Theta_K$ is given by

\bea
\label{eq:mob_pdf}
f_{\und{\Theta}}(\underline{\theta}) =
%\begin{cases}
    B(\underline{\alpha}, \underline{\beta})^{-1} \prod_{i=1}^K \theta_i^{\alpha_i-1}(1-\theta_i)^{\beta_i-1}, & \und{\theta} \in S,  
   % 0, & \text{otherwise},
%\end{cases}
\eea
where $\un{\theta}=(\theta_1, \cdots, \theta_K)$, and
\bea
\label{eq:monotone_set}
S \coloneqq \{ \un{\theta} \in [0,1]^K \mid \theta_1 \ge \theta_2 \ge \cdots \ge \theta_K\}
\eea is the set of vectors that satisfy the monotonicity constraint, and $B(\und{\alpha}, \und{\beta})$ is the normalization constant. 
%. Furthermore, $B(\und{\alpha}, \und{\beta})$ is the normalization factor given by: %that ensures that $\int_{\theta \in S} f_{\un{\Theta}}(\theta)=1$. %given by
% \iffalse
% \bean
% \scalebox{0.98}{$
% B(\underline{\alpha}, \underline{\beta}) \coloneqq \int\limits_{0}^{1} \int\limits_{\theta_K}^{1} \cdots \int\limits_{\theta_2}^{1} \left( \prod\limits_{i=1}^K \theta_i^{\alpha_i-1} \, (1-\theta_i)^{\beta_i-1} \right)   \textup{d}\theta_1 \cdots \textup{d}\theta_K.$}
% \eean
% \fi 
% \[
% B(\underline{\alpha}, \underline{\beta}) = \int\limits_{0}^{1} \int\limits_{\theta_K}^{1} \cdots \int\limits_{\theta_2}^{1} \left( \prod\limits_{i=1}^K \theta_i^{\alpha_i-1} \, (1-\theta_i)^{\beta_i-1} \right)   \textup{d}\theta_1 \cdots \textup{d}\theta_K.
% \]
\begin{figure*}[ht!]
\centering
\begin{subfigure}{.3\textwidth}
  \centering
  \includegraphics[width=0.9\textwidth]{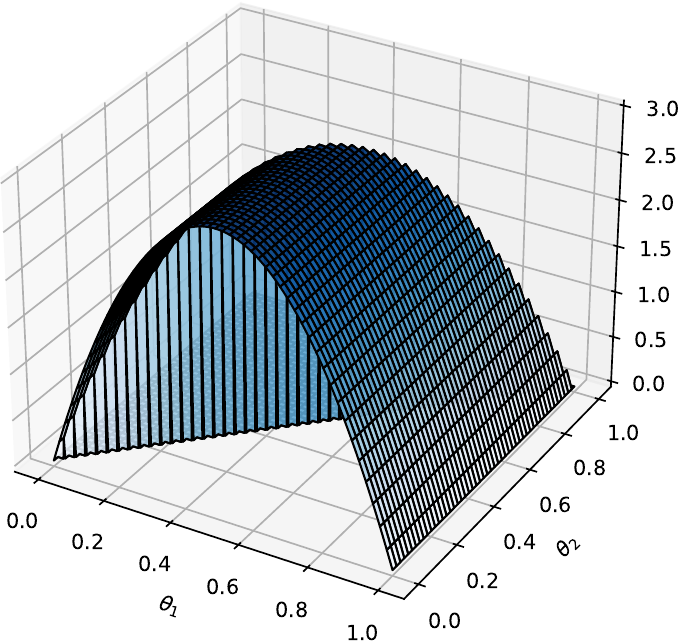}
  \caption{Joint PDF of the $\mathrm{MOB}$ distribution.}
  \label{fig:mob_pdf}
\end{subfigure}%
\hfill
\begin{subfigure}{.3\textwidth}
  \centering
  \includegraphics[width=0.9\textwidth]{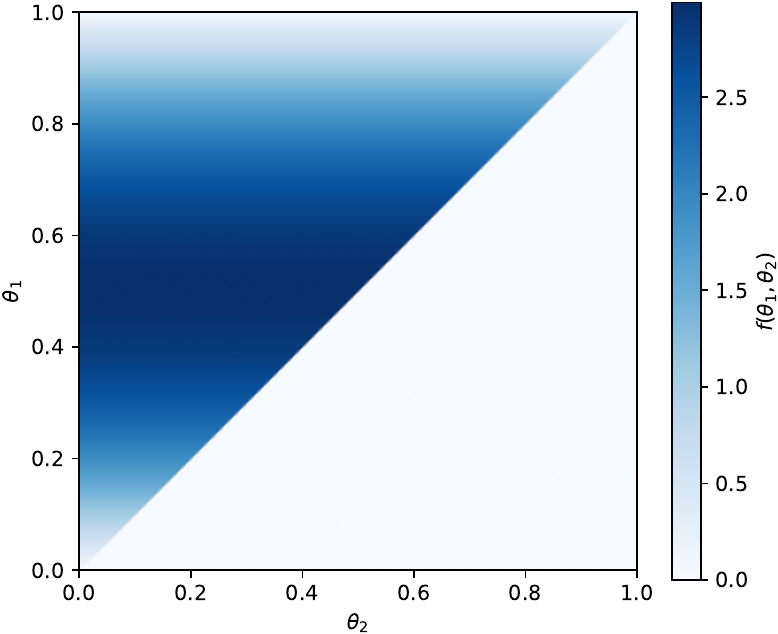}
  \caption{Heat map of $\mathrm{MOB}$ distribution. }% The PDF is nonzero in the triangle $0 \le \theta_2 \le \theta_1 \le 1$ and shows the effect of the ordering constraint}
  \end{subfigure}
  \hfill
\begin{subfigure}{.3\textwidth}
  \centering
  \includegraphics[width=0.9\textwidth]{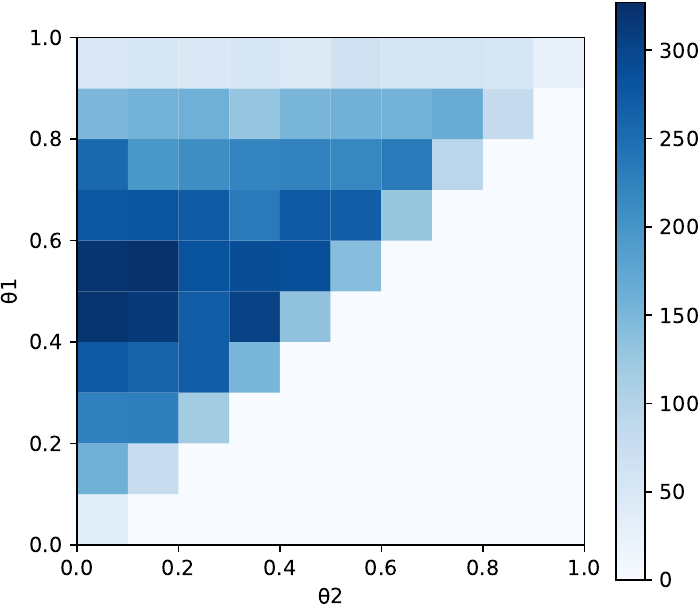}
  \caption{Histogram of $10000$ $\mathrm{MOB}$ samples generated via Gibbs sampling approach. 
  %that is run for $N=1000$ steps. %We can see that the samples have histogram similar to that of the PDF shown in figure~\ref{fig:mob_pdf}. %{\color{red} remove the title on top of this figure, not very visible, please replace the figures with pdfs instead of PNG} 
  }
  \label{fig:gibbs_out}
\end{subfigure}
\caption{Illustrations of the PDF of $\mathrm{MOB}((2,1),(2,1))$ distribution and the histogram samples collected using Gibbs sampling. The PDF is nonzero in the triangle $0 \le \theta_2 \le \theta_1 \le 1$ and shows the effect of the ordering constraint.}
\label{fig:mob_pdf_histogram}
\end{figure*}

% \begin{figure*}[ht!]
% \centering
% \includegraphics[width=0.8\linewidth]{uniform.png}
% \caption{Joint PDF of the Multivariate Ordered Beta Distribution with $K=2$, $\alpha_i=1$, $\beta_i=1$. Density is nonzero in the triangle $0 < \theta_2 < \theta_1 < 1$ and shows the effect of the ordering constraint. {\color{red} MV:remove the title on the top, remove largest next to theta1 label}}
% \label{fig:joint_pdf_k2}
% \end{figure*}

\subsection{Parameter Updates in the Joint-TS Algorithm}
%We will now show that the update step under the Joint-TS algorithm remains identical to that under the TS algorithm. 
Let $X_{i,t}$ denote the reward from arm $i$ at time $t$. Then,
\bean
P(X_{i, t} = x \mid \Theta_i = \theta) ~=~ \theta^x (1-\theta)^{1-x}, \quad x \in \{0,1\}.
\eean
On playing arm $i$ and observing reward $X_{i,t} = r \in \{0,1\}$, the posterior on the parameters $\underline{\Theta} = (\Theta_1, \cdots, \Theta_K)$ is given by: 
\bean
\label{eq:posterior-MOB}
f_{\underline{\Theta} | X_{i,t}}(\underline{\theta} \mid r)
    \propto  f_{\Theta}(\underline{\theta}) 
    \cdot P(X_{i, t} = r \mid \Theta_i = \theta)\\
    \propto 
    \scalebox{0.8}{$\left(\displaystyle\prod_{j=1}^K \theta_j^{\alpha_j-1}(1-\theta_j)^{\beta_j-1} \right) \theta_i^{r} (1-\theta_i)^{1 - r}$}, & \un{\theta} \in S
\eean
Clearly,~\eqref{eq:posterior-MOB} represents an MOB distribution with parameters
\begin{equation}
(\alpha_i, \beta_i) \leftarrow (\alpha_i + r, ~~\beta_i+1-r), \quad (\alpha_j, \beta_j) \leftarrow (\alpha_j, \beta_j) ~ \forall j \neq i.
\label{eq:parameter-updates}
\end{equation}
Though the parameters of arms $j \ne i$ remain unchanged, the reward from arm $i$ implicitly acts as feedback for the other arms due to the monotonicity constraint. 

Unlike classical TS, which samples independent Beta distributions, Joint-TS, as seen in Algorithm~\ref{alg:jointts_pc}, samples the vector $(\Theta_1, \ldots, \Theta_K)$ jointly at each step to inherently satisfy $\theta_1 \geq \theta_2 \geq \dots \geq \theta_K$. This implicit correlation makes direct sampling from the MOB distribution a non-trivial task, motivating our Gibbs sampling approach outlined below.
%Noting that MOB distributions  introduced very recently in \cite{OrderedBeta_SaidiKuznetsovNediak} and due to the fact that the random variables, $\Theta_1, \cdots, \Theta_K$ are correlated, sampling from this distribution is non-trivial. 
% Below, we outline a Gibbs sampling approach to obtain samples from the MOB distribution.

\begin{algorithm}[ht!]
\caption{Joint-Thompson Sampling}
\label{alg:jointts_pc}
\begin{algorithmic}[1]
\State \textbf{Initialize:} For all $i \in [K]$, choose $\alpha_i > 0$, $\beta_i > 0$.
\For{$t = 1, 2, \ldots, T$}
    %\State \textcolor{gray}{// \emph{Joint parameter sampling step}}
    \State Sample $(\Theta_1, \ldots, \Theta_K) \sim \operatorname{MOB} (\underline{\alpha}, \underline{\beta})$ (Algorithm~\ref{alg:Gibbs-sampling}).
    \State Select arm $I_t = \arg\max_{i} \Theta_i s_i$
    \State Observe reward $R_t=r$ from arm $I_t$
    \State $(\alpha_{I_t}, \ \beta_{I_t}) \leftarrow (\alpha_{I_t}+r, \ \beta_{I_t} + 1-r)$
\EndFor
\end{algorithmic}
\end{algorithm}
\bnote
If $\alpha_i=a$ and $\beta_i=b$ for all $i \in [K]$, for some $a,b>0$, MOB samples can be generated by sorting $K$ i.i.d Beta$(a, b)$ samples. This idea does not extend to general MOB distributions where $\alpha_i$ and $\beta_i$ can take any values $>0$.
\enote

\section{Sampling From MOB Distribution}
We use the Gibbs sampling technique \cite{geman1984stochastic} to sample from MOB distribution. Gibbs Sampling is an MCMC technique to sample from complex joint distributions that are difficult to directly sample from, but whose conditional distributions are easy to sample from. In what follows, we first define a {\em restricted Beta distribution}, and subsequently demonstrate that if a random vector is MOB distributed, then the conditional distributions follow restricted Beta distribution.

%\subsection{Restricted Beta distribution}
\bdefn
A random variable $Y$ is said to follow {\em restricted Beta distribution} with parameters $(\alpha, \beta, \ell, u)$, denoted by $Y \sim {\rm ResBeta}(\alpha, \beta, \ell, u)$, if the PDF of $Y$ is given by
\bea
\label{eq:restr_beta_distr}
f_{Y}(y) = \begin{cases}
\frac{y^{\alpha - 1} (1-y)^{\beta - 1}}{B(\alpha, \beta, \ell, u)}, & \quad y \in [\ell, u], \\
0, & \quad \text{otherwise},
\end{cases}
\eea
where it is assumed that $\alpha, \beta > 0$, $0 \le \ell < u \le 1$, and $B(\alpha, \beta, \ell, u)$ is the normalization factor.
%\bea
%\label{eq:beta_partial}
% B(\alpha, \beta, \ell, u) = \int_{\ell}^{u} y^{\alpha-1} \ (1-y)^{\beta-1} \ \textup{d}y.
% \eea
\edefn
If $\ell=0$ and $u=1$, then ${\rm ResBeta}(\alpha, \beta, 0, 1) = {\rm Beta}(\alpha, \beta)$.

\subsection{Sampling from Restricted Beta Distribution}
Let $Y \sim \mathrm{ResBeta}(a, b, \ell, u)$ and let $X \sim \mathrm{Beta}(a, b)$. %From~\eqref{eq:restr_beta_distr}~and~\eqref{eq:beta_partial}, 
It follows that the CDF of $Y$ is given by
\bean
F_Y(y) %&=& (F_X(y)-F_X(\ell))*\frac{B(a, b)}{B(a, b, \ell, u)}\\
&=& \frac{F_X(y)-F_X(\ell)}{F_X(u)-F_X(\ell)}, \qquad y \in [\ell, u].
\eean
The well-known inverse transform sampling method may be used to sample from restricted Beta distribution.
%Since Python libraries are available to access CDF and inverse CDF of Beta distributions, we can use inverse transform sampling method to sample from restricted Beta distribution. %as presented in the code snippet below.
% The below snippet of Python code demonstrates how to generate a sample from the restricted Beta distribution.

% \begin{lstlisting}[caption={Python code to generate a sample from the restricted Beta distribution.},numbers=none]
% import numpy as np
% from scipy.stats import Beta as betad

% def generate_restricted_beta(alpha, Beta, lb, ub):
%     uni_sample = np.random.uniform()
%     LB = betad.cdf(lb, alpha, Beta)
%     UB = betad.cdf(ub, alpha, Beta)

%     return betad.ppf(LB + uni_sample * (UB - LB), alpha, Beta)
% \end{lstlisting}

% From Lemma~\ref{lem:cond_distr} of $\Theta_i$ conditioned over $\un{\Theta}_{\sim i} = \un{\theta}_{\sim i}$ is $\mathrm{ResBeta}(\alpha_i, \beta_i, \theta_{i+1}, \theta_{i-1})$. We will show how one can sample from the restricted Beta distribution using the stats libraries available in Python to access its CDF and inverse CDF functions.
\subsection{Conditional Distribution $f_{\Theta_i \mid \un{\Theta}_{\sim i}}$ under MOB} \label{sec:sampling_mob} 

We now proceed to show that if a random vector is MOB distributed, then the conditional PDF of each random variable in the vector, conditioned over the remaining random variables, follows restricted Beta distribution.

% \begin{figure}[ht!]
% \centering
% \includegraphics[width=0.4\textwidth]{static_mcs.pdf}
% \caption{MCS values picked by the OLLA, TS, Joint-TS algorithms.}
% \end{figure}

\blem \label{lem:cond_distr}
If $\un{\Theta} \sim \mathrm{MOB}(\un{\alpha}, \un{\beta})$, then
\[
\Theta_i \mid \{\un{\Theta}_{\sim i} = \un{\theta}_{\sim i}\} \sim \mathrm{ResBeta}(\alpha_i, \beta_i, \theta_{i+1}, \theta_{i-1}),
\]
where by convention, $\theta_{0}=1$, $\theta_{K+1}=0$.
% conditional density of $\Theta_i$ given $\Theta_j = \theta_j$ for all $j \in [K] \setminus \{i\}$ is given by:
% \bean
% f_{\Theta_i \mid \Theta_{\sim i}}(\theta_i \mid \theta_{\sim i}) = \begin{cases}
% \frac{\theta_i^{\alpha_i-1}(1-\theta_i)^{\beta_i-1}}{B(\alpha_i, \beta_i, \theta_{i+1}, \theta_{i-1})} & \theta_i \in [\theta_{i+1}, \theta_{i-1}]\\
% 0 & \text{otherwise}
% \end{cases}
% \eean
% where $\un{\Theta}_{\sim i} = (\Theta_1, \cdots, \Theta_{i-1}, \Theta_{i+1}, \cdots, \Theta_K)$, similarly $\un{\theta}_{\sim i} = (\theta_1, \cdots, \theta_{i-1}, \theta_{i+1}, \cdots, \theta_K)$, $\theta_{-1}=1$, $\theta_{K+1}=0$ and $B(\alpha_i, \beta_i, \theta_{i+1}, \theta_{i-1})$ is the normalization factor given by:
% \bea
% \label{eq:partial_beta}
% B(\alpha_i, \beta_i, \theta_{i+1}, \theta_{i-1}) = \int_{\theta_{i+1}}^{\theta_{i-1}} \theta_i^{\alpha_i-1}(1-\theta_i)^{\beta_i-1} d\theta_i.
% \eea
\elem
\bprf
Follows directly from the joint PDF in~\eqref{eq:mob_pdf}.
% \bean
% &f_{\Theta_i \mid \un{\Theta}_{\sim i}}(\theta_i \mid \un{\theta}_{\sim i}) = \frac{f_{\un{\Theta}}(\un{\theta})} {f_{\un{\Theta}_{\sim i}}(\un{\theta}_{\sim i})}\\
% &\propto 
% \begin{cases} \theta_i^{\alpha_i-1} (1-\theta_i)^{\beta_i-1}, & 
% \theta_i \in [\theta_{i+1},\theta_{i-1}], \\
% 0, & \text{otherwise}.
% \end{cases}
% \eean
% The proportionality above follows from the definition of the MOB distribution in~\eqref{eq:mob_pdf}. The proportionality constant is a function of $\un{\theta}_{\sim i}$. Because the conditional PDF integrates to $1$, i.e., $\int_{\theta_{i+1}}^{\theta_i} f_{\Theta_i \mid \un{\Theta}_{\sim i}}(\theta_i \mid \un{\theta}_{\sim i}) d\theta_i = 1$, it follows from~\eqref{eq:beta_partial} that the proportionality constant is given by $1/B(\alpha_i, \beta_i, \theta_{i+1}, \theta_{i-1})$.
\eprf

% It is clear to see that the conditioned over $\Theta_{\sim i} = \theta_{\sim i}$, $\Theta_i$ is $\mathrm{ResBeta}(\alpha, \beta, \theta_{i+1}, \theta_{i-1})$.

\subsection{Gibbs Sampling from MOB Distribution}
\label{subsec:Gibbs-sampling}

Fix $N \in \mathbb{N}$ and $(\und{\alpha}, \und{\beta})$ with $\alpha_i>0$ and $\beta_i>0$ for all $i \in [K]$. Let $\un{\Theta}_1, \un{\Theta}_2, \ldots$ be sampled as per the Gibbs sampling procedure outlined in Algorithm~\ref{alg:Gibbs-sampling}.

\begin{algorithm}[h]
\caption{Gibbs Sampling from ${\rm MOB}(\und{\alpha}, \und{\beta})$ Distribution}
\label{alg:Gibbs-sampling}
\begin{algorithmic}[1]

\State {\bf Input:} $(\und{\alpha}, \und{\beta})$ with $\alpha_i>0$ and $\beta_i>0$ for all $i \in [K]$, number of iterations $N \in \mathbb{N}$. 

\State Generate $U_1, \ldots, U_K \stackrel{\rm i.i.d.}{\sim} \mathrm{Uniform}([0,1])$, and order the samples samples in descending order. Let the ordered samples be $\Theta_{1,1}, \cdots, \Theta_{1, K}$, with $\Theta_{1,1} \ge \Theta_{1,2} \ge \cdots \ge \Theta_{1, K}$. Let $\un{\Theta}_1 = (\Theta_{1,1}, \cdots, \Theta_{1, K})$.

\For{$t = 2, \ldots, N$}
\For{$i = 1, \ldots, K$}
    \State Sample $\Theta_{t,i} \sim \mathrm{ResBeta}(\alpha_i, \beta_i, \Theta_{t-1, i+1}, \Theta_{t, i-1})$ for all $i \in [K]$; by convention, $\Theta_{t,0}=1$ and $\Theta_{t,K+1}=0$.
\EndFor
 \State Set $\und{\Theta}_t = (\Theta_{t,1}, \ldots, \Theta_{t,K})$.
\EndFor

\State {\bf Output:} $\und{\Theta}_{N} = (\Theta_{N,1}, \ldots, \Theta_{N,K})$.
\end{algorithmic}
\end{algorithm}

\blem
\label{lem:ergodicity-properties-of-Gibbs-sampling}
The sequence of random vectors $\un{\Theta}_1, \un{\Theta}_2, \cdots $ obtained via Algorithm~\ref{alg:Gibbs-sampling} forms an $S$-valued Markov chain (where $S$ is as defined in \eqref{eq:monotone_set}). Furthermore, 
%Furthermore, the preceding Markov chain is $\lambda$-invariant (where $\lambda$ is the Lebesgue measure on $\mathbb{R}^K)$ and aperiodic, with ${\rm MOB}(\und{\alpha}, \und{\beta})$ as its invariant distribution.
%\elem
%The proof of the Lemma is omitted as it merely involves a routine verification of the general conditions of invariance of aperiodicity for general state Markov chains; see \cite{tierney1994markov} or \cite[Chapter~13]{meyn2012markov} for the exact definitions of invariance and aperiodicity for general state Markov chains.
%Using Lemma~\ref{lem:ergodicity-properties-of-Gibbs-sampling} in conjunction with \cite[Theorem~1]{tierney1994markov}, we deduce that the Markov chain $\{\und{\Theta}_1, \und{\Theta}_2, \ldots\}$ is positive Harris-recurrent, and
$$
\lim_{t \to \infty} \ \|\mathcal{L}(\und{\Theta}_t \mid \und{\Theta}_1=\und{\theta}) - {\rm MOB}(\und{\alpha}, \und{\beta})\|_{\rm TV} = 0 \qquad \forall ~ \und{\theta} \in S,
$$
where $\mathcal{L}(\und{\Theta}_t \mid \und{\Theta}_1=\und{\theta})$ denotes the conditional law of $\und{\Theta_t}$, conditioned on starting the Gibbs sampling in state $\und{\Theta}_1=\und{\theta}$, and $\|\cdot\|_{\rm TV}$ denotes the total variation distance.
\elem
We skip the proof of this Lemma. It follows from the fact that the Markov chain $\{\und{\Theta}_1, \und{\Theta}_2, \ldots\}$ can be shown to be positive Harris-recurrent \cite{tierney1994markov}, \cite[Chapter~13]{meyn2012markov}. 
%{\color{red} Karthik: I have cut this short, please check if this line is still meaningful}.

Lemma~\ref{lem:ergodicity-properties-of-Gibbs-sampling} assures that for $N$ sufficiently large, the joint PDF of the samples output by Algorithm~\ref{alg:Gibbs-sampling} is close to the desired ${\rm MOB}(\und{\alpha},\und{\beta})$ distribution in total variation. In Joint-TS, we run Algorithm~\ref{alg:Gibbs-sampling} with $N=1000$ at each time step. For the case with $K=2$ and $\un{\alpha}=(2,1)$, $\un{\beta}=(2,1)$, 
%{\color{blue} Karthik: Is it possible to re-run the simulation with a different choice of $\und{\alpha}$ and $\und{\beta}$? The current choice corresponds to the simple case of uniform distribution, and reviewers might not see the usefulness of the Gibbs sampling approach for this simple case} 
it can be seen in Fig.~\ref{fig:gibbs_out} that the 2D histogram of the samples generated via Algorithm~\ref{alg:Gibbs-sampling} matches with the PDF of MOB distribution.

% {\color{red} To Karthik: Can we get a reference for the above statement ?, can we get any supporting theory for finite runs of Markov chain to approximate the distribution ?

\bnote
The sequential inverse transform sampling method in \cite{conTS} suggests sampling $\Theta_1 \sim \mathrm{Beta}(\alpha_1, \beta_1)$ and $\Theta_i \mid \und{\Theta}_{\sim i} \sim \mathrm{ResBeta}(\alpha_2, \beta_2, 0, \Theta_{i-1})$ for all $i > 1$. This fundamentally differs from the true conditional distributions of the MOB (see Sec.~\ref{sec:sampling_mob}). For example, for $K=2$, $\underline{\alpha} = (1000, 1000)$ and $\underline{\beta} = (1000, 1)$, the MOB joint PDF is concentrated around $(1,1)$, whereas the method in \cite{conTS} falsely concentrates $\Theta_1$ around 0.5. %{\color{blue} Karthik: we are commenting about the joint PDF under MOB on one hand and on the marginal PDF of $\Theta_1$ on the other; more like comparing apples to oranges. For consistency, can we just say explicitly what will be the joint PDF using the technique of \cite{conTS}, or what will be the marginal PDF of $\Theta_1$ under MOB?}. 
\enote

% betsequentially samples each of $(\Theta_1, \ldots, \Theta_K)$ from Beta$(\alpha, \beta)$ while constraining each sample to be smaller than the previous one. However, this procedure does not necessarily comply with the true marginal distributions of the success probabilities. For example, in a two-arm bandit $(\Theta_0, \Theta_1)$ with $\underline{\alpha} = (1000, 1000)$ and $\underline{\beta} = (1000, 1)$, the joint pdf is concentrated near $(1,1)$. But under sequential sampling, $\Theta_0$ would instead be sampled from a distribution concentrated around $0.5$. 
% \enote

\section{Simulation Setup and Results} \label{sec:simulations}
The simulations are carried out using the Python itpp repository developed by Saxena et. al \cite{py_itpp} that provides libraries to simulate the fading and Doppler models. We consider $K = 29$ MCS options available in 5G NR (Table~\ref{tab:MCS_Table_5G}) and evaluate Joint-TS against OLLA, Modified Thompson Sampling (MTS) \cite{MTS}, Unimodal Thompson Sampling (UTS) \cite{UTS}, and Latent-TS (LTS)\cite{Vidit_LatentTS}. Note that LTS requires a pre-existing mapping table to correlate latent states with MCS success probabilities. While we assume this table is perfectly accurate in our simulations, relying on static lookup tables in real-world deployments can cause sub-optimal performance when actual channel conditions deviate from the modeled mapping.
% While the first setting enables us to compare the convergence rate of the TS and Joint-TS algorithms with that of OLLA, the second setting models realistic doppler scenarios. % Static channel results help understand the performance of the aforementioned algorithms in the absence of CQI. 
% %We assume that OLLA has perfect CQI and hence assume its performance is the best possible. 
% We do not, however, compare with other LA approaches such as Latent Thompson Sampling~\cite{Vidit_LatentTS}, where SINR value is learned and mapped to an MCS value through existing tables, or upper confidence bound (UCB)-based LA~\cite{UCB}. 
% Though the MCS values are $29$, the CQI feedback provided by the user equipment is quantized to following MCS values
% {\color{red} add CQI values and corresponding reference}

\begin{figure*}[ht!]
\centering
\begin{subfigure}{.285\textwidth}
  \centering
  \includegraphics[width=\textwidth]{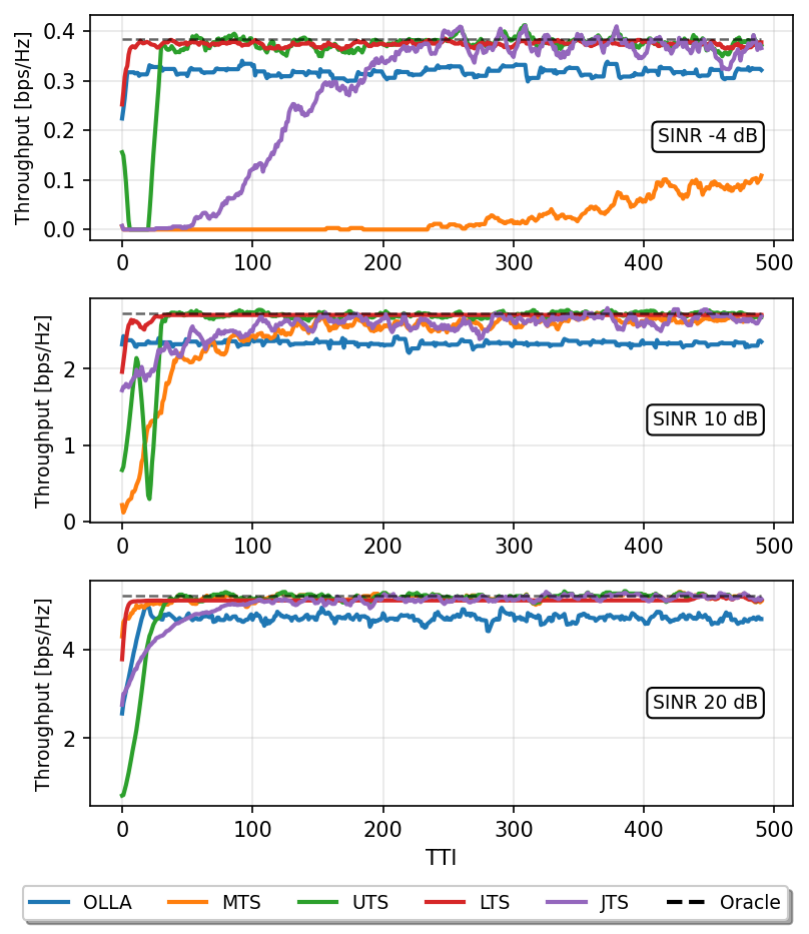}
  \caption{Static channel setting}
  \label{fig:static_tput}
\end{subfigure}%
 \hfill
\begin{subfigure}{.285\textwidth}
  \centering
  \includegraphics[width=\textwidth]{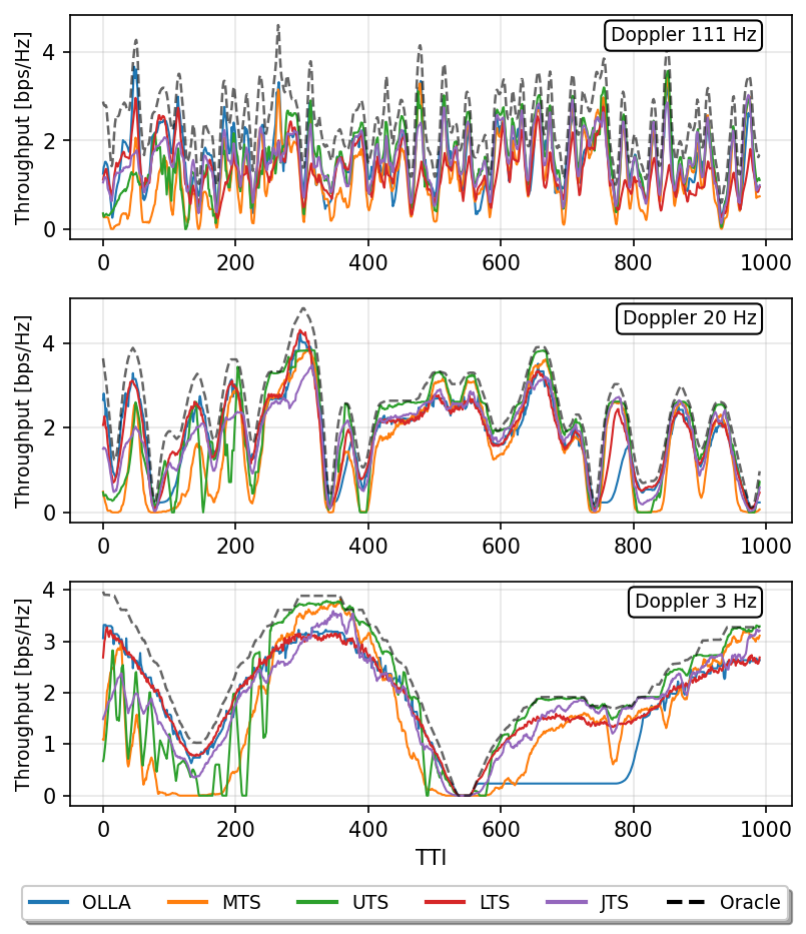}
  \caption{Perfect CQI at average SINR 10dB}
  \label{fig:doppler-tput-10}
\end{subfigure}
\hfill
\begin{subfigure}{.285\textwidth}
  \centering
  \includegraphics[width=\textwidth]{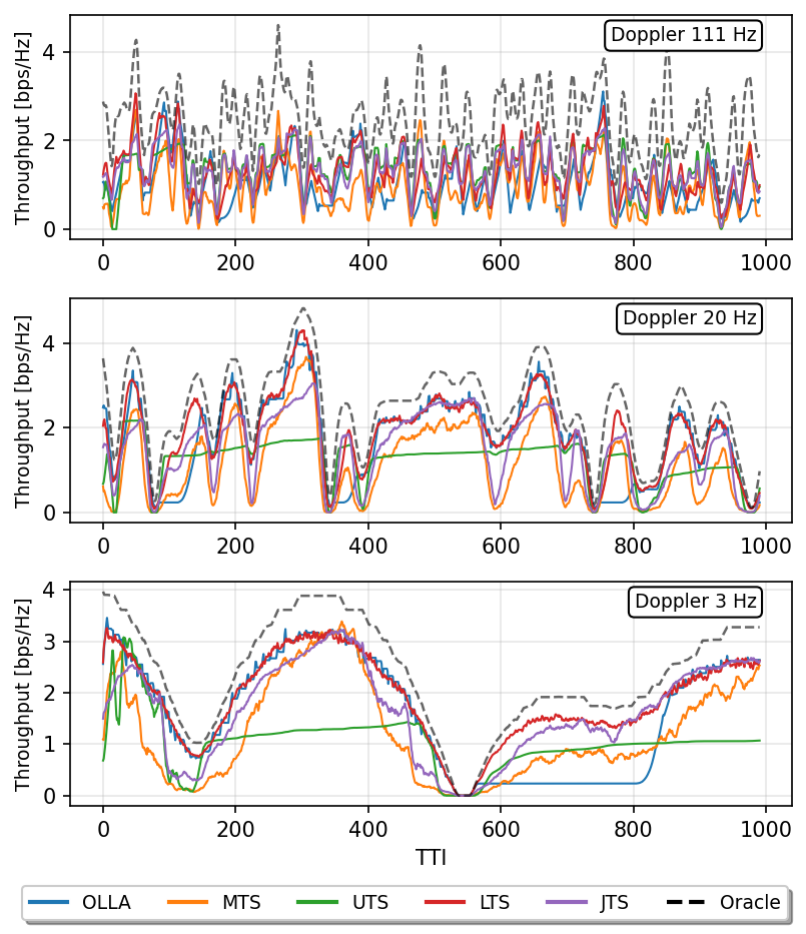}
  \caption{Without CQI at average SINR 10dB}
  \label{fig:doppler_no_cqi}
\end{subfigure}
\caption{Throughput of OLLA, MTS, UTS, LTS, Joint-TS algorithms averaged over 20 samples. Oracle in the last 2 figures implies correspond to best possible throughput for the channel conditions. It also indicates the effect of Doppler on the SINRs.}
\label{fig:sim_results}
\end{figure*}

\paragraph{ Static Channel Simulations}
We initialize $\alpha_i=\beta_i=1$ for all $i \in [K]$ to run the TS and Joint-TS algorithms at three different SINR values: -4dB, 10dB, 20dB. Averaged over 20 independent trials, stable conditions allow OLLA, UTS, and LTS to exhibit the highest aggregate throughput. Joint-TS demonstrates competitive performance (e.g., achieving 2.54 bps/Hz at 10 dB SINR compared to LTS's 2.68 bps/Hz).
\vspace{4pt}
\paragraph{ Doppler Channel} We present simulations for the Doppler setting in Fig.~\ref{fig:doppler-tput-10} and ~\ref{fig:doppler_no_cqi} that correspond to scenarios with perfect CQI and CQI-less settings. We assume unconstrained optimization to find the optimal arm for both these simulations. %~and~\ref{fig:doppler-tput-15}. 
Three cases of Doppler shifts, $F_d=3, 20, 111$ Hz, are considered at average SINR 10dB. This corresponds to coherence times of $\approx 60,9,1.6$ ms respectively, using the relation $T_{\text{coh}}=\frac{9}{16\pi F_d }$. We have assumed a single-path channel model here.  
% \paragraph{Perfect CQI setting} We assume that CQI is available for each TTI (500$\mu$s) and run 15 parallel multi-armed bandit instances, each corresponding to a particular CQI. Note that for those SINRs whose best MCS values are one of $18, 19, 20$, the CQI is set to $18$, as in \cite[5.2.2.1]{3gpp38_214}, courtesy of MCS quantization. So even though the CQI is assumed available, the MAB corresponding to a single CQI can potentially observe a range of SINRs over which single MCS need not be optimal. It can be seen that both TS and Joint-TS adapt to changes in the channel. The throughput gains from using Joint-TS algorithm over TS are significantly visible at lower rates when SINR drops. As the time progresses, the TS algorithm eventually adapts to achieve similar rates as Joint-TS. % For SINR of 15dB, TS is observed to dominate initially, but Joint-TS eventually adapts to achieve similar rates as TS. 
%Fig.~\ref{fig:doppler_tput}.
% \begin{figure}[ht!]
% \includegraphics[width=0.5\textwidth]{doppler_throughput_SINR10.pdf}
% \end{figure}
\begin{table}[ht!]
    \centering
    \scalebox{0.7}{\begin{tabular}{|c|c|c|c|c|c|c|}
    \hline
    \multirow{2}{*}{CQI setting} & \multirow{2}{*}{\shortstack{Doppler\\Shift (Hz)}} & \multicolumn{5}{|c|}{Throughput (in bps/Hz)}\\ \cline{3-7}
    & & OLLA & MTS & UTS & LTS & Joint-TS\\ \hline
    \multirow{3}{*}{Perfect CQI}  & 3 & 1.64 & 1.53 & 1.89 & 1.89 & 1.80 \\ \cline{2-7}
         & 20 & 1.81  &  1.56 &   1.89 &   1.87  &  1.80 \\ \cline{2-7}
         & 111 & 1.61 &  1.24 &   1.56  &  1.29  &  1.56 \\ \hline
    \multirow{3}{*}{No CQI}  & 3 & 1.60  & 1.24  &  1.04 &    1.88  & 1.63 \\ \cline{2-7}
         & 20 & 1.72  &  1.13   & 1.21   & 1.87  &  1.49 \\ \cline{2-7}
         & 111 &  1.06  & 0.96  &  1.34  &  1.29 &   1.32 \\ \hline
    \end{tabular}}
    \caption{Average throughput across the 1000 TTIs at average SINR of 10db.}
    \label{tab:placeholder}
\end{table}

\underline{\textit{Perfect CQI setting}}: Assuming error and delay free CQI is available every TTI (500 $\mu$s), we run 20 parallel MAB instances. Note that for SINRs optimally mapped to $18, 19, 20$ are grouped to CQI mapped to MCS $18$, as in \cite[5.2.2.1]{3gpp38_214}, due to MCS quantization. Therefore, even though the CQI is assumed available, each MAB may observe a range of SINRs over which single MCS need not be optimal. %It can be seen that both TS and Joint-TS adapt to changes in the channel. The throughput gains from using Joint-TS over TS are significantly visible at lower rates when SINR drops. Over time, the TS algorithm eventually adapts to achieve similar rates as Joint-TS.%
While UTS and LTS achieve the highest throughput at low/mid Doppler shifts, LTS degrades significantly at high Doppler shifts, whereas UTS and Joint-TS successfully maintain high throughput.

\underline{\textit{CQI-less settings}}: When CQI feedback is unavailable, a single MAB instance of a smoothed MTS/UTS/Joint-TS variant is used. The parameter update is modified to:
$(\alpha_{I_t}, \beta_{I_t}) \leftarrow (\alpha_{I_t}, \beta_{I_t})\, e^{-\frac{\Delta t}{w}} + (r,\, 1-r),$
where $\Delta t$ denotes the time interval since arm $I_t$ was last explored, and $w$ is a smoothing parameter \cite{wifi}. A higher value of $w$ corresponds to longer memory and slower adaptation to change, while a lower value of $w$ results in faster response.  Fig.~\ref{fig:doppler_no_cqi} shows results for $w=50$. The parameter $w$ can be adapted based on Doppler estimates to achieve optimal performance. A detailed analysis of this is left for future work. We use the smoothening method proposed in \cite{Vidit_LatentTS} for LTS. In this setting, UTS struggles without CQI, while LTS maintains high throughput in low-mid Doppler settings. Joint-TS demonstrates consistently robust performance across all conditions.

% algorithms with faster convergence rate have better throughput performance. At SINRs 10dB and 20dB, we observe that Joint-TS converges after processing feedback from 100 packets. If the channel coherence time is larger than time taken for convergence, these gains are significant. However, for the cellular setting, this would imply a coherence time of $\gg 100*500\mu {\rm s} = 50 {\rm ms}$.

% \begin{table}[ht!]
%     \centering
%     \begin{tabular}{||c|c|c|c|c||}
%     \hline
%     \multirow{2}{*}{Average SINR} & \multirow{2}{*}{Doppler Shift (Hz)} & \multicolumn{3}{c||}{Throughput (in bps/Hz)}\\ \cline{3-5}
%     & & OLLA & TS & Joint-TS \\ \hline
%     \multirow{3}{*}{15db}  & 3 & 3.13 & 2.81 & 2.97\\ \cline{2-5}
%          & 20 & 3.24 & 2.87 & 2.95 \\ \cline{2-5}
%          & 111 & 2.79 & 2.34 & 2.57 \\ \hline
%     \end{tabular}
%     \caption{Average throughput across the entire experiment}
%     \label{tab:placeholder}
% \end{table}

% \begin{table}[ht!]
%     \centering
%     \begin{tabular}{||c|c|c|c|c||}
%     \hline
%     \multirow{2}{*}{Average SINR} & \multirow{2}{*}{Doppler Shift (Hz)} & \multicolumn{3}{c||}{Throughput (in bps/Hz)}\\ \cline{3-5}
%     & & OLLA & TS & Joint-TS \\ \hline
%     \multirow{3}{*}{20db}  & 3 & 4.25 & 4.15 & 4.08\\ \cline{2-5}
%          & 20 & 4.25 & 4.14 & 4.08 \\ \cline{2-5}
%          & 111 & 3.83 & 3.61 & 3.66 \\ \hline
%     \end{tabular}
%     \caption{Average throughput across the entire experiment}
%     \label{tab:placeholder}
% \end{table}

\section{Conclusion}
Joint-TS exploits the monotonicity property of success probabilities of MCS values by generating samples from MOB distribution, thereby resulting in faster convergence in low to mid-range SINRs and throughput gains in both static and Doppler settings with and without CQI. Future work will investigate the performance of Joint-TS in scenarios with periodic (infrequent) CQI feedback and under conditions of noisy or imperfect CQI estimation.

% {\color{red}
% \section{To-Do (following up on WCNC review comments)}
% \begin{enumerate}
%     \item Add all the implementations to collab (static settings)
%     \item Convergence guarantees for finite $N$, i.e., to be within $\epsilon$-distance from MOB distribution, how many samples do we need?
%     (Karthik to check with Sarath about convergence guarantees).

%     \item Throughput comparison with UTS, Latent TS, G-ORS, (check what LinConTS is).  

%     \item Compare with CoTS (srikant)
%     \item Compare ordered Thompson Sampling with jointTS
%     \item Check constrained UCB with vanilla TS

%     \item Averaging over more than $10$ trials. And check the confidence intervals. \textbf{}

%     \item Runtime of inverse cdf function, check results with N less than 1000 (500, 250, etc)
% \end{enumerate}
% }

\bibliographystyle{IEEEtran}
\bibliography{LinkAdapt_spcomm.bib}

\begin{thebibliography}{10}
\providecommand{\url}[1]{#1}
\csname url@samestyle\endcsname
\providecommand{\newblock}{\relax}
\providecommand{\bibinfo}[2]{#2}
\providecommand{\BIBentrySTDinterwordspacing}{\spaceskip=0pt\relax}
\providecommand{\BIBentryALTinterwordstretchfactor}{4}
\providecommand{\BIBentryALTinterwordspacing}{\spaceskip=\fontdimen2\font plus
\BIBentryALTinterwordstretchfactor\fontdimen3\font minus
  \fontdimen4\font\relax}
\providecommand{\BIBforeignlanguage}[2]{{%
\expandafter\ifx\csname l@#1\endcsname\relax
\typeout{** WARNING: IEEEtran.bst: No hyphenation pattern has been}%
\typeout{** loaded for the language `#1'. Using the pattern for}%
\typeout{** the default language instead.}%
\else
\language=\csname l@#1\endcsname
\fi
#2}}
\providecommand{\BIBdecl}{\relax}
\BIBdecl

\bibitem{eff_SINR}
K.~Brueninghaus, D.~Astely, T.~Salzer, S.~Visuri, A.~Alexiou, S.~Karger, and
  G.-A. Seraji, ``{Link Performance Models for System Level Simulations of
  Broadband Radio Access Systems},'' in \emph{IEEE 16th International Symposium
  on Personal, Indoor and Mobile Radio Communications (PIMRC)}, vol.~4, 2005,
  pp. 2306--2311 Vol. 4.

\bibitem{bler_target}
P.~Wu and N.~Jindal, ``{Coding Versus ARQ in Fading Channels: How Reliable
  Should the PHY Be?}'' in \emph{IEEE Global Communications Conference
  (GLOBECOM)}, 2009, pp. 1--6.

\bibitem{optimal_target_bler}
S.~Park, R.~C. Daniels, and R.~W. Heath, ``{Optimizing the Target Error Rate
  for Link Adaptation},'' in \emph{IEEE Global Communications Conference
  (GLOBECOM)}, 2015, pp. 1--6.

\bibitem{3gpp38_214}
{3rd Generation Partnership Project (3GPP)}, ``{NR; Physical layer procedures
  for data},'' 3GPP, Technical Specification TS 38.214, 2018, release 15.

\bibitem{OLLA_3G}
D.~Paranchych and M.~Yavuz, ``{A Method for Outer Loop Rate Control in High
  Data Rate Wireless Networks},'' in \emph{Proceedings IEEE 56th Vehicular
  Technology Conference (VTC)}, vol.~3, 2002, pp. 1701--1705 vol.3.

\bibitem{OLLA_LTE}
K.~I. Pedersen, G.~Monghal, I.~Z. Kovacs, T.~E. Kolding, A.~Pokhariyal,
  F.~Frederiksen, and P.~Mogensen, ``{Frequency Domain Scheduling for OFDMA
  with Limited and Noisy Channel Feedback},'' in \emph{IEEE 66th Vehicular
  Technology Conference (VTC)}, 2007, pp. 1792--1796.

\bibitem{OLLA_shortcomings}
V.~Buenestado, J.~M. Ruiz-Avilés, M.~Toril, S.~Luna-Ramírez, and A.~Mendo,
  ``{Analysis of Throughput Performance Statistics for Benchmarking LTE
  Networks},'' \emph{IEEE Communications Letters}, vol.~18, no.~9, pp.
  1607--1610, 2014.

\bibitem{OLLA_self_opt}
A.~Durán, M.~Toril, F.~Ruiz, and A.~Mendo, ``{Self-Optimization Algorithm for
  Outer Loop Link Adaptation in LTE},'' \emph{IEEE Communications Letters},
  vol.~19, no.~11, pp. 2005--2008, 2015.

\bibitem{UCB}
R.~Combes, J.~Ok, A.~Proutiere, D.~Yun, and Y.~Yi, ``{Optimal Rate Sampling in
  802.11 Systems: Theory, Design, and Implementation},'' \emph{IEEE
  Transactions on Mobile Computing}, vol.~18, no.~5, pp. 1145--1158, 2019.

\bibitem{vidit_epsilon}
V.~Saxena, J.~Jald\'{e}n, J.~E. Gonzalez, M.~Bengtsson, H.~Tullberg, and
  I.~Stoica, ``{Contextual Multi-Armed Bandits for Link Adaptation in Cellular
  Networks},'' in \emph{Proceedings of the 2019 Workshop on Network Meets AI \&
  ML}, ser. NetAI'19.\hskip 1em plus 0.5em minus 0.4em\relax New York, NY, USA:
  Association for Computing Machinery, 2019, p. 44–49.

\bibitem{MTS}
H.~Gupta, A.~Eryilmaz, and R.~Srikant, ``{Low-Complexity, Low-Regret Link Rate
  Selection in Rapidly-Varying Wireless Channels},'' in \emph{IEEE INFOCOM 2018
  - IEEE Conference on Computer Communications}, 2018, pp. 540--548.

\bibitem{Vidit_TS}
V.~Saxena and J.~Jaldén, ``{Bayesian Link Adaptation under a BLER Target},''
  in \emph{IEEE 21st International Workshop on Signal Processing Advances in
  Wireless Communications (SPAWC)}, 2020, pp. 1--5.

\bibitem{UTS}
S.~Paladino, F.~Trov\`{o}, M.~Restelli, and N.~Gatti, ``{Unimodal thompson
  sampling for graph-structured arms},'' in \emph{Proceedings of the
  Thirty-First AAAI Conference on Artificial Intelligence}, ser. AAAI'17.\hskip
  1em plus 0.5em minus 0.4em\relax AAAI Press, 2017, p. 2457–2463.

\bibitem{Vidit_LatentTS}
V.~Saxena, H.~Tullberg, and J.~Jaldén, ``{Reinforcement Learning for Efficient
  and Tuning-Free Link Adaptation},'' \emph{IEEE Transactions on Wireless
  Communications}, vol.~21, no.~2, pp. 768--780, 2022.

\bibitem{wifi}
A.~Krotov, A.~Kiryanov, and E.~Khorov, ``{Rate Control With Spatial Reuse for
  Wi-Fi 6 Dense Deployments},'' \emph{IEEE Access}, vol.~8, pp.
  168\,898--168\,909, 2020.

\bibitem{OrderedBeta_SaidiKuznetsovNediak}
M.~Al-Saidi, A.~Kuznetsov, and M.~Nediak, ``{On Ordered Beta Distribution and
  the Generalized Incomplete Beta Function},'' \emph{Method. Comput. Appl.
  Prob.}, vol.~27, no.~1, Dec. 2024.

\bibitem{py_itpp}
V.~Saxena, ``{py-itpp},'' \url{https://github.com/vidits-kth/py-itpp},
  accessed: 2025-09-13.

\bibitem{conTS}
H.~Gupta, A.~Eryilmaz, and R.~Srikant, ``{Link Rate Selection using Constrained
  Thompson Sampling},'' in \emph{IEEE INFOCOM 2019 - IEEE Conference on
  Computer Communications}, 2019, pp. 739--747.

\bibitem{thompson1933likelihood}
W.~R. Thompson, ``{On the Likelihood That One Unknown Probability Exceeds
  Another in View of the Evidence of Two Samples},'' \emph{Biometrika},
  vol.~25, no. 3/4, pp. 285--294, 1933.

\bibitem{geman1984stochastic}
S.~Geman and D.~Geman, ``{Stochastic Relaxation, {G}ibbs Distributions, and the
  {B}ayesian Restoration of Images},'' \emph{IEEE Transactions on pattern
  analysis and machine intelligence}, no.~6, pp. 721--741, 1984.

\bibitem{tierney1994markov}
L.~Tierney, ``{Markov Chains for Exploring Posterior Distributions},''
  \emph{the Annals of Statistics}, pp. 1701--1728, 1994.

\bibitem{meyn2012markov}
S.~P. Meyn and R.~L. Tweedie, \emph{{Markov Chains and Stochastic
  Stability}}.\hskip 1em plus 0.5em minus 0.4em\relax Springer Science \&
  Business Media, 2012.

\end{thebibliography}

\end{document}